# Learning a Target Sample Re-Generator for Cross-Database Micro-Expression Recognition

Yuan Zong[*]
Key Laboratory of Child Development and Learning Science of Ministry of Education, Research Center for Learning Science, Southeast University
Nanjing 210096, China
xhzongyuan@seu.edu.cn

Xiaohua Huang
Center for Machine Vision and Signal Analysis, Faculty of Information Technology and Electrical Engineering, University of Oulu
Oulu FI-90014, Finland
xiaohua.huang@oulu.fi

Wenming Zheng[†]
Key Laboratory of Child Development and Learning Science of Ministry of Education, Research Center for Learning Science, Southeast University
Nanjing 210096, China
wenming_zheng@seu.edu.cn

Zhen Cui
School of Computer Science and Engineering, Nanjing University of Science and Technology
Nanjing 210094, China
zhen.cui@njust.edu.cn

Guoying Zhao
Center for Machine Vision and Signal Analysis, Faculty of Information Technology and Electrical Engineering, University of Oulu
Oulu FI-90014, Finland
guoying.zhao@oulu.fi

## ABSTRACT

In this paper, we investigate the cross-database micro-expression recognition problem, where the training and testing samples are from two different micro-expression databases. Under this setting, the training and testing samples would have different feature distributions and hence the performance of most existing micro-expression recognition methods may decrease greatly. To solve this problem, we propose a simple yet effective method called Target Sample Re-Generator (TSRG) in this paper. By using TSRG, we are able to re-generate the samples from target micro-expression database and the re-generated target samples would share same or similar feature distributions with the original source samples. For this reason, we can then use the classifier learned based on the labeled source samples to accurately predict the micro-expression categories of the unlabeled target samples. To evaluate the performance of the proposed TSRG method, extensive cross-database micro-expression recognition experiments designed based on SMIC and CASME II databases are conducted. Compared with recent state-of-the-art cross-database emotion recognition methods, the proposed TSRG achieves more promising results.

[*]Yuan Zong is also with the Center for Machine Vision and Signal Analysis, Faculty of Information Technology and Electrical Engineering, University of Oulu, Oulu FI-90014, Finland.
[†]Corresponding author.

## 1 INTRODUCTION

Micro-expression is one type of particular facial expressions and it can reveal human beings' true emotional states which people try to conceal [29]. Therefore, recognizing micro-expressions by machines will have many valuable applications, e.g., clinical diagnosis [10], interrogation [11], and security [27]. However, compared with ordinary facial expression, micro-expression has much lower intensity and shorter duration. This makes automatic micro-expression recognition become a very challenging task. Nevertheless, micro-expression recognition is still one of recent attractive research topics among affective computing, multimedia information processing and pattern recognition communities [26] due to its potential values.

The micro-expression recognition research can be early traced to the work of [29], in which Pfister et al. proposed to use temporal interpolation model (TIM) and local binary pattern from three orthogonal planes (LBP-TOP) [44] to deal with micro-expression recognition problem. Their experimental results show that LBP-TOP is effective for micro-expression recognition problem. Following Pfister et al.'s work, Ruiz-Hernandez et al. [30] employed re-parameterization of second order Gaussian jet to boost LBP-TOP such that LBP-TOP is more applicable to micro-expression recognition. For better describing micro-expressions, Wang et al. [39] proposed a novel spatio-temporal descriptor called LBP with six intersection points (LBP-SIP) which can reduce the redundant information in LBP-TOP. Subsequently, lots of spatio-temporal descriptors are developed for micro-expression recognition tasks, such as spatio-temporal LBP with integration projection (STLBP-IP) [15], completed local quantized pattern-TOP (CLQP-TOP) [16], histogram of oriented gradient-TOP (HOG-TOP) [22], and histogram of image gradient orientation-TOP (HIGO-TOP) [22]. Furthermore, different from the above spatio-temporal descriptors, other types of micro-expression features are also investigated by researchers. Among them, it is worth mentioning the works of [25, 41], in which Liu et al. and Xu et al. respectively designed novel and effective features, i.e., main directional mean optical (MDMO) and facial dynamics map (FDM), to describe micro-expressions. Their experimental results demonstrated the effectiveness of these two novel micro-expression features.

On the other hand, in recent years, some researchers investigated micro-expression recognition from a new angle, i.e., aiming at leveraging other important information of micro-expression clips, which contributes to distinguishing micro-expressions, to boost the performance of the spatio-temporal descriptors. In the work of [38], Wang et al. proposed to use robust principal component analysis (RPCA) [40] to extract the background information from the micro-expression video clips and then extract the spatio-temporal descriptor of the background information to describe such micro-expression clips. Recently, Wang et al. [36, 37] designed a set of regions of interest (ROIs) according to the facial action coding system (FACS) [9] for micro-expression feature extraction. Meanwhile, they proposed a color space decomposition method called tensor independent color space (TICS) to utilize the color information for micro-expression recognition. It is notable that before Wang et al.'s ROIs based method [36, 37], most researchers employed the fixed gird based spatial division method, e.g., 8 × 8, to boost the performance of spatio-temporal descriptors. Specifically, the original the micro-expression video clip is first divided into a set of spatial facial blocks and then the spatio-temporal descriptors are extracted to compose a supervector to describe the micro-expressions. More recently, deep learning methods have also been applied to micro-expression recognition. Kim et al. [20] proposed a deep learning framework consisting of popular convolutional neural network (CNN) [21] and long short-term memory (LSTM) recurrent network [12] for micro-expression recognition. In this framework, the representative expression-states frames of micro-expression video clips are first selected to train a CNN. Then, the CNN feature of each image frame in a video clip is extracted to train a LSTM network for recognizing micro-expressions.

Although micro-expression recognition has made great progress in recent years, it should be pointed out that nearly all of the above proposed methods are just considered to evaluate on one micro-expression database, which means the training and testing samples belong to the same micro-expression database. In this case, since the training and testing samples are collected by the same equipment and under the same environment, it is a common view that the training and testing samples abide by the same or similar feature distributions. However, in practical applications, we will face the micro-expression samples recorded by different equipments or under different environments, which inevitably brings the feature distribution difference between the training and testing samples. Because of this, the performance of existing micro-expression recognition methods may sharply drop. Consequently, in order to develop more practical micro-expression recognition methods, it is very worthwhile to investigate cross-database micro-expression problem, in which the training and testing samples come from two different micro-expression databases. Clearly, cross-database micro-expression recognition problem is more challenging and difficult than ordinary micro-expression recognition one. For convenience, we refer the training database (samples) as the source database (samples) and the testing database (samples) as the target database (samples) in cross-database micro-expression recognition problem throughout this paper.

In the work of [42], Yan et al. roughly divided cross-database facial expression recognition problem, which is closely related to our topic, into two cases including semi-supervised case and unsupervised case. The major difference between these two cases is whether we have access to the label information of target domain. Similarly, cross-database micro-expression recognition problem can follow this categorization. In this paper, we will focus on the unsupervised problem setting, in which the source micro-expression samples are labeled while the label information of the target micro-expression samples is completely unknown. To deal with this challenging problem, we propose a simple yet effective method called Target Sample Re-Generator (TSRG). TSRG aims at learning a sample re-generator for source and target micro-expression samples. When the source and target samples are fed to TSRG, respectively, the output of TSRG for source samples will still be themselves, while for target samples TSRG will output a new set of samples which are different from their original forms but shares same or similar feature distributions with the source sample set. After that, we are able to learn a classifier such as support vector machine (SVM) based on the labeled source micro-expression samples and subsequently use it to predict the labels of the re-generated target micro-expression samples.

The rest of this paper is organized as follows: Section 2 reviews recent cross-database emotion (including facial expression and speech emotion) recognition works which are very closely related to cross-database micro-expression recognition topic. In Section 3, we introduce our proposed TSRG based cross-database micro-expression recognition method in detail. For evaluating the performance of the proposed TSRG method, extensive cross-database micro-expression recognition experiments between SMIC and CASME II databases are conducted in Section 4. Finally, the paper is concluded in Section 5.

## 2 RELATED WORKS

Since cross-database micro-expression recognition problem has not yet been investigated, in this section we review recent works about other modality based cross-database emotion recognition that is closely related to cross-database micro-expression recognition including cross-database facial expression recognition and cross-database speech emotion recognition. In recent years, these two challenging and interesting problems have gained lots of researchers' attention.

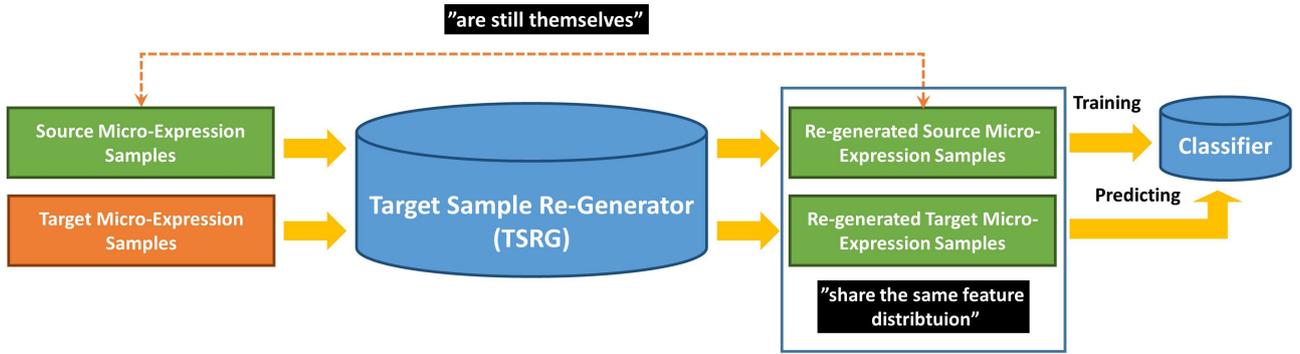

Figure 1: The overall picture of TSRG based cross-database micro-expression recognition method.

For cross-database facial expression recognition and its related problems, various effective methods [3, 4, 32, 42, 46, 47] have been proposed. For example, in the works of [3, 4], Chu et al. proposed a novel method called selective transfer machine (STM) for personalized (cross-subject) facial action units detection problem. STM is able to utilize the target samples to learn a set of weights for the source samples such that the weighted source samples would have the same or similar feature distributions with the target samples. Consequently, the classifier trained based on the weighted source samples could also be suitable for distinguishing the target samples. Recently, Sangineto et al. [32] investigated the cross-domain facial expression recognition problem by using a transductive parameter transfer method. They proposed a novel classifier parameter transfer method to directly transfer knowledge about the parameters of source person-specific classifiers to the target individuals such that the target classifier can accurately predict the expressions of target samples. More recently, Yan et al. [42] proposed an unsupervised domain-adaptive dictionary learning (UDADL) model to cope with the unsupervised cross-database facial expression recognition problem and achieved a promising result. In addition, Zheng et al. [46, 47] proposed a transductive transfer subspace learning framework to deal with cross-pose and cross-database cases in facial expression recognition. In this framework, an auxiliary set is selected from the unlabeled target samples for learning a subspace together with the labeled source samples. In such a subspace, the source and target samples would be enforced to abide by the same feature distribution and hence the classifier trained by the source samples can predict the expressions of the target samples.

The earliest cross-database speech emotion recognition research may be the work of [33], in which Schuller et al. proposed to employ a series of normalization methods to investigate cross-database speech emotion recognition problem and conducted extensive cross-database experiments on many speech emotion databases. From then on, a variety of interesting methods have been proposed to deal with this challenging problem [5–7, 13, 17, 31, 34, 48]. For example, Hassan et al. [13] proposed an importance-weighted support vector machine (IW-SVM) to handle cross-database speech emotion recognition. In this method, they first used three transfer learning [28] methods, i.e., kernel mean matching (KMM) [14], Kullback-Leibler importance estimation procedure (KLIEP) [35], and unconstrained least-squares importance fitting (uLSIF) [19] to learn the importance weights for target samples with respect to source samples and then incorporated the learned weights into SVM to obtain the IW-SVM. In the works of [5–7], Deng et al. proposed a series of auto-encoder based domain adaptation methods by leveraging various auto-encoder based networks to learn a common representation between the source and target samples for cross-database speech emotion recognition problem. Recently, Song et al. [34] proposed a transfer non-negative matrix factorization method for coping with cross-database speech emotion recognition, in which the maximum mean discrepancy (MMD) [1] is introduced to eliminate the feature distribution difference between source and target speech databases.

From the above cross-database emotion recognition methods, it is clear that the basic idea of these methods can be almost categorized into two types. The first type of methods target at learning the importance weights for source or target samples to balance the feature distribution difference between source and target databases, while the second type of methods achieve this goal by learning a common subspace. Different from the above methods, our proposed TSRG is designed to this end from a new angle, i.e., re-generating target samples in the original feature space.

## 3 PROPOSED METHOD
### 3.1 Overall Picture and Basic Idea

For better understanding TSRG method, we draw a picture shown in Fig. 1 to illustrate the basic idea of the TSRG method and show how TSRG works for dealing with cross-database micro-expression recognition problem. As depicted in Fig. 1, it can be seen that the goal of the proposed TSRG is to learn a sample re-generator which can re-generate the source and target micro-expression samples by inputting the original source and target ones. Interestingly, the re-generated source micro-expression samples are still themselves, while the re-generated target micro-expression samples are different from their original forms but their feature distribution becomes same or similar with the source samples. Consequently, once the optimal TSRG is learned, we can train a classifier, e.g., SVM, based on the labeled source samples and then obtain the micro-expression categories of the unlabeled target samples by using the trained classifier to predict the labels of the corresponding new target samples re-generated by TSRG.

### 3.2 Building a Target Sample Re-Gegerator

Suppose we have feature matrices $\mathbf{X}_s \in \mathbb{R}^{d \times n_s}$ and $\mathbf{X}_t \in \mathbb{R}^{d \times n_t}$ of source and target micro-expression samples from two different databases, where $d$ is the dimension of feature vectors and $n_s$ and $n_t$ denote the numbers of source samples and target samples, respectively. Note that the feature here can be any widely-used micro-expression feature such as LBP-TOP [29, 44], LBP-SIP [39], and MDMO [25]. To obtain the functions of the sample re-generator shown in Fig. 1, firstly, the sample re-generator must output the source samples themselves with the source samples as input, which can be formulated as the following optimization problem:

$$\min_{G} \|\mathbf{X}_s - G(\mathbf{X}_s)\|_F^2, \quad (1)$$

where $G$ denotes the sample re-generator to be learned and $F$ denotes the Frobenius norm.

Secondly, to ensure that the re-generated target samples have same or similar feature distributions with the source samples, we should also design a function $f_G(\mathbf{X}_s, \mathbf{X}_t)$ for TSRG, whose details will be introduced in what follows. By using $f_G(\mathbf{X}_s, \mathbf{X}_t)$ to serve as the regularization term, we can obtain the optimization problem of TSRG as follows:

$$\min_{G} \|\mathbf{X}_s - G(\mathbf{X}_s)\|_F^2 + \lambda f_G(\mathbf{X}_s, \mathbf{X}_t), \quad (2)$$

where $\lambda$ is the trade-off parameter to control the balance between these two terms of the objective function.

It is notable that the output $G(\mathbf{X}_s)$ of TSRG for the source samples is hoped to be still the original source samples $\mathbf{X}_s$. This goal is actually easy to achieve by a combination of the kernel mapping operation and the linear projection operation, which are two typical operations in subspace learning. More specifically, a suitable sample re-generator $G$ can first map the source samples from the original feature space into a reproduced kernel Hilbert space (RKHS) by a kernel mapping operator $\phi$ and subsequently transform the infinite-dimensional source features in RKHS back to the original feature space by a projection matrix $\phi(\mathbf{C}) \in \mathbb{R}^{\infty \times d}$. Following this idea, the sample re-generator $G$ can be finally defined as $G(\cdot) = \phi(\mathbf{C})^T \phi(\cdot)$. Then the optimization problem of TSRG in Eq. (2) can be written as:

$$\min_{\phi(\mathbf{C})} \|\mathbf{X}_s - \phi(\mathbf{C})^T \phi(\mathbf{X}_s)\|_F^2 + \lambda f_G(\mathbf{X}_s, \mathbf{X}_t). \quad (3)$$

As seen from Eq. (3), the sample re-generator $G$ consisting of kernel mapping and linear projection operators can re-generate $\mathbf{X}_s$ themselves. More importantly, it can also bring a benefit to construct $f_G(\mathbf{X}_s, \mathbf{X}_t)$ for TSRG. It is known that we are able to eliminate the feature distribution difference between two different feature sets by minimizing their maximum mean discrepancy (MMD) [1] which is defined in a RKHS. Therefore, regarding $f_G(\mathbf{X}_s, \mathbf{X}_t)$ of our TSRG, we can formulate it as the MMD between source and target samples in the RKHS produced by $\phi$, which is expressed as:

$$\mathrm{MMD}(\mathbf{X}_s, \mathbf{X}_t) = \|\frac{1}{n_s}\phi(\mathbf{X}_s)\mathbf{1}_s - \frac{1}{n_t}\phi(\mathbf{X}_t)\mathbf{1}_t\|_{\mathcal{H}}, \quad (4)$$

where $\mathcal{H}$ denotes a Hilbert space, $\mathbf{1}_s$ and $\mathbf{1}_t$ are the vectors with the lengths of $n_s$ and $n_t$, respectively, and their elements are all one. However, it is hard to directly learn the optimal kernel mapping operator $\phi$. Therefore, we relax MMD in Eq. (4) to the following formulation to serve as $f_G(\mathbf{X}_s, \mathbf{X}_t)$ for TSRG such that we only need to learn the optimal $\phi(\mathbf{C})$, which is feasible and also consistent with the model parameter of TSRG to be learned in Eq. (3):

$$f_G(\mathbf{X}_s, \mathbf{X}_t) = \|\frac{1}{n_s}\phi(\mathbf{C})^T\phi(\mathbf{X}_s)\mathbf{1}_s - \frac{1}{n_t}\phi(\mathbf{C})^T\phi(\mathbf{X}_t)\mathbf{1}_t\|_2^2. \quad (5)$$

We have following lemma to show that minimizing MMD in Eq. (4) is equivalent to minimizing the proposed $f_G(\mathbf{X}_s, \mathbf{X}_t)$ in Eq. (5), which can support our relaxation.

LEMMA 3.1. *For* $\mathrm{MMD}(\mathbf{X}_s, \mathbf{X}_t)$ *and* $f_G(\mathbf{X}_s, \mathbf{X}_t)$ *defined as Eqs. (4) and (5) based on the kernel mapping operator $\phi$, we have* $f_G(\mathbf{X}_s, \mathbf{X}_t) \to 0$ *if* $\mathrm{MMD}(\mathbf{X}_s, \mathbf{X}_t) \to 0$.

PROOF. From the condition that $\mathrm{MMD}(\mathbf{X}_s, \mathbf{X}_t)$ is close to 0, we know that $\phi(\mathbf{X}_s)$ and $\phi(\mathbf{X}_t)$ have similar expectations, which can be formulated as $\mathbb{E}(\phi(\mathbf{x}_s^i)) \approx \mathbb{E}(\phi(\mathbf{x}_t^i))$, where $\mathbb{E}(\cdot)$ denotes the expectation operator. Then according to the linear property of expectation [2], i.e., $\mathbb{E}(Ax) = A\mathbb{E}(x)$, it is easy to deduce that $\mathbb{E}(\phi(\mathbf{C})^T\phi(\mathbf{x}_s^i)) \approx \mathbb{E}(\phi(\mathbf{C})^T\phi(\mathbf{x}_t^i))$, which guarantees $f_G(\mathbf{X}_s, \mathbf{X}_t)$ will be close to 0. □

By substituting the proposed $f_G(\mathbf{X}_s, \mathbf{X}_t)$ in Eq. (5) into TSRG in Eq. (3), the optimization problem of TSRG becomes as follows:

$$\min_{\phi(\mathbf{C})} \|\mathbf{X}_s - \phi(\mathbf{C})^T\phi(\mathbf{X}_s)\|_F^2$$
$$+\lambda\|\frac{1}{n_s}\phi(\mathbf{C})^T\phi(\mathbf{X}_s)\mathbf{1}_s - \frac{1}{n_t}\phi(\mathbf{C})^T\phi(\mathbf{X}_t)\mathbf{1}_t\|_2^2. \quad (6)$$

To solve TSRG, let $\phi(\mathbf{C}) = [\phi(\mathbf{X}_s), \phi(\mathbf{X}_t)]\mathbf{P}$, where $\mathbf{P} \in \mathbb{R}^{(n_s+n_t) \times d}$. Then by using the kernel trick, the optimization problem of TSRG can be converted to the following formulation:

$$\min_{\mathbf{P}} \|\mathbf{X}_s - \mathbf{P}^T\mathbf{K}_s\|_F^2 + \lambda\|\frac{1}{n_s}\mathbf{P}^T\mathbf{K}_s\mathbf{1}_s - \frac{1}{n_t}\mathbf{P}^T\mathbf{K}_t\mathbf{1}_t\|_2^2 + \mu\|\mathbf{P}\|_1, \quad (7)$$

where $\mathbf{K}_s = [\mathbf{K}_{ss}^T, \mathbf{K}_{st}^T]^T$ and $\mathbf{K}_t = [\mathbf{K}_{ts}^T, \mathbf{K}_{tt}^T]^T$. Herein $\mathbf{K}_{ss} = \phi(\mathbf{X}_s)^T\phi(\mathbf{X}_s)$, $\mathbf{K}_{st} = \phi(\mathbf{X}_s)^T\phi(\mathbf{X}_t)$, $\mathbf{K}_{ts} = \phi(\mathbf{X}_t)^T\phi(\mathbf{X}_s)$ and $\mathbf{K}_{tt} = \phi(\mathbf{X}_t)^T\phi(\mathbf{X}_t)$, and they can be computed by different kernel functions such as Gaussian kernel. This is the final formulation of the proposed TSRG. Note that in Eq. (7), we also introduce $l_1$ norm with respect to $\mathbf{P}$, i.e., $\|\mathbf{P}\|_1 = \sum_{i=1}^{d}\|\mathbf{p}_i\|_1$, where $\mathbf{p}_i$ is the $i$-th column of $\mathbf{P}$, for TSRG to serve as the regularization term. There are two important reasons. Firstly, it can avoid the overfitting problem [8] during optimizing TSRG. Secondly, each column of $\phi(\mathbf{C})$ will be enforced to reconstruct by all the columns of $\phi(\mathbf{X}_s)$ and $\phi(\mathbf{X}_t)$ sparsely, which is more reasonable. The sparsity of $\mathbf{P}$ is controlled by the trade-off parameter $\mu$.

## 3.3 Optimization

The optimization problem of TSRG can be easily solved by various methods such as iterative thresholding (IT) [40], accelerated proximal gradient (APG) [18], exact augmented Lagrange multiplier (EALM) [24] and inexact ALM (IALM) [24]. In this paper, we employ IALM method to learn the optimal $\mathbf{P}$ of TSRG. More specifically, we introduce a new variable $\mathbf{Q}$ which equals $\mathbf{P}$ for TSRG and then convert TSRG optimization problem of Eq. (7) to a constrained one as follows:

$$\min_{\mathbf{P},\mathbf{Q}} \|\mathbf{X}_s - \mathbf{Q}^T \mathbf{K}_s\|_F^2 + \lambda \|\frac{1}{n_s}\mathbf{Q}^T \mathbf{K}_s \mathbf{1}_s - \frac{1}{n_t}\mathbf{Q}^T \mathbf{K}_t \mathbf{1}_t\|_F^2 + \mu \|\mathbf{P}\|_1,$$
$$\text{s.t. } \mathbf{P} = \mathbf{Q}. \quad (8)$$

Subsequently, the Lagrange function can be obtained as the following formulation:

$$L(\mathbf{P},\mathbf{Q},\mathbf{T},\kappa) = \|\mathbf{X}_s - \mathbf{Q}^T \mathbf{K}_s\|_F^2 + \lambda \|\mathbf{Q}^T \Delta \mathbf{k}_{st}\|_2^2 + \mu \|\mathbf{P}\|_1$$
$$+ tr[\mathbf{T}^T(\mathbf{P}-\mathbf{Q})] + \frac{\kappa}{2}\|\mathbf{P}-\mathbf{Q}\|_F^2, \quad (9)$$

where $\mathbf{T}$ is the Lagrange multiplier, $\kappa > 0$ is a regularization parameter, and $\Delta \mathbf{k}_{st} = \frac{1}{n_s}\mathbf{K}_s \mathbf{1}_s - \frac{1}{n_t}\mathbf{K}_t \mathbf{1}_t$.

Finally, to learn the optimal $\mathbf{P}$, we only need to minimize the Lagrange function in Eq. (9) with respect to one of the variables while fixing the others iteratively. Specifically, repeating the following four steps until obtaining convergence:
(1) Fix $\mathbf{P}$, $\mathbf{T}$, $\kappa$ and update $\mathbf{Q}$:

In this step, the optimization problem with respect to $\mathbf{Q}$ is as follows:

$$\min_{\mathbf{Q}} \|\mathbf{X}_s - \mathbf{Q}^T \mathbf{K}_s\|^2 + \lambda \|\mathbf{Q}^T \Delta \mathbf{k}_{st}\|_F^2 + tr[\mathbf{T}^T(\mathbf{P}-\mathbf{Q})]$$
$$+ \frac{\kappa}{2}\|\mathbf{P}-\mathbf{Q}\|_F^2,$$

whose close-form solution is

$$\mathbf{Q} = (\mathbf{K}_s \mathbf{K}_s^T + \Delta \mathbf{k}_{st} \Delta \mathbf{k}_{st}^T + \frac{\kappa \mathbf{I}}{2})^{-1}(\mathbf{K}_s \mathbf{X}_s^T + \frac{\kappa \mathbf{P} + \mathbf{T}}{2}).$$

(2) Fix $\mathbf{Q}$, $\mathbf{T}$, $\kappa$ and update $\mathbf{P}$:

$$\min_{\mathbf{P}} \frac{\mu}{\kappa}\|\mathbf{P}\|_1 + \frac{1}{2}\|\mathbf{P}-(\mathbf{Q}-\frac{\mathbf{T}}{\kappa})\|_F^2.$$

By using the soft-thresholding operator, we are able to obtain the optimal $\mathbf{P}$ according to the following criterion:

$$\mathbf{P}_{ij} = \begin{cases} (\mathbf{Q}_{ij} - \frac{\mathbf{T}_{ij}}{\kappa}) - \frac{\mu}{\kappa}, & \text{if } (\mathbf{Q}_{ij} - \frac{\mathbf{T}_{ij}}{\kappa}) > \frac{\mu}{\kappa}; \\ (\mathbf{Q}_{ij} - \frac{\mathbf{T}_{ij}}{\kappa}) + \frac{\mu}{\kappa}, & \text{if } (\mathbf{Q}_{ij} - \frac{\mathbf{T}_{ij}}{\kappa}) < \frac{\mu}{\kappa}; \\ 0, & \text{otherwise}. \end{cases}$$

where $\mathbf{P}_{ij}$, $\mathbf{Q}_{ij}$, and $\mathbf{T}_{ij}$ are the elements in the $i^{th}$ row and $j^{th}$ column of their corresponding matrices.
(3) Update $\mathbf{T}$ and $\kappa$:

$$\mathbf{T} = \mathbf{T} + \kappa(\mathbf{P}-\mathbf{Q}), \quad \kappa = \min(\rho\kappa, \kappa_{max}),$$

where $\rho$ is a scaled parameter.
(4) Check convergence:

$$\|\mathbf{P}-\mathbf{Q}\|_\infty < \epsilon,$$

where $\epsilon$ denotes the machine epsilon.

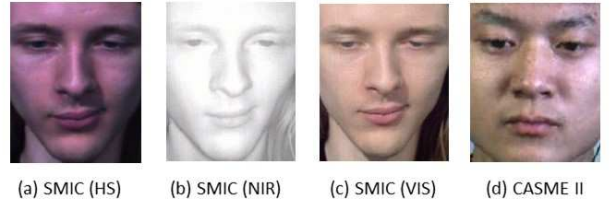

Figure 2: Examples of SIMC and CASME II micro-expression databases. From left to right, they are image frames of the video clips from (a) SMIC (HS), (b) SMIC (NIR), (c) SMIC (VIS), and (d) CASME II, respectively.

## 4 EXPERIMENTS

### 4.1 Micro-Expression Databases and Experiment Protocol

In this section, we conduct extensive experiments between SMIC and CASME II micro-expression databases to evaluate our proposed TSRG based cross-database micro-expression recognition method. SMIC database[1] is built by Li et al. [23]. It has three types of datasets, i.e., SMIC (HS), SMIC (VIS), and SMIC (NIR), which are recorded by a high speed (HS) camera of 100fps, a normal visual (VIS) camera of 25fps, and a near-infrared (NIR) camera, respectively. SMIC (HS) contains 164 micro-expression clips from 16 different subjects, while SMIC (VIS) and SMIC (NIR) both consist of 71 samples from 8 participants. The samples of three datasets of SMIC are all divided into three micro-expression categories, i.e., *Positive*, *Negative*, and *Surprise*. CASME II database[2] is collected by Yan et al. [43]. It includes 26 subjects and records their 247 micro-expression samples. These samples are categorized into five micro-expression classes including *Happiness*, *Surprise*, *Disgust*, *Repression*, and *Others*, respectively. In this paper, the face images in the video clips from CASME II database are cropped and then transformed to 308 × 257 pixels, while for the samples from three SMIC databases, we crop and transform the images into 170 × 139 pixels for experiments. To see the difference among three datasets of SMIC and CASME II, we select an image frame from the micro-expression video clip belonging to four datasets, respectively, which are shown in Fig. 2 as the examples.

The cross-database micro-expression experiments in this paper are between CASME II and one dataset of SMIC, i.e., CASME II v.s. SMIC (HS), CASME II v.s. SMIC (VIS), and CASME II v.s. SMIC (NIR). The two datasets in each above combination are alternatively served as source and target databases. Therefore, there will be six groups of experiments in total. For convenience, we denote these six experiments by Exp.1, Exp.2, Exp.3, Exp.4, Exp.5, and Exp.6, respectively. In order to make CASME II and three datasets of SMIC share the same micro-expression categorization, we select the samples belonging to *Happiness*, *Surprise*, *Disgust*, and *Repression* from CASME II and re-label these samples. Specifically, the samples of *Happiness* are given *Positive* labels and the samples of *Disgust* and

---
[1]SMIC database can be freely obtained from http://www.cse.oulu.fi/SMICDatabase.
[2]CASME II database is publicly available and can be downloaded from http://fu.psych.ac.cn/CASME/casme2-en.php.

**Table 1: The numbers of different micro-expression samples in CASME II and SMIC datasets for cross-database micro-expression recognition experiments.**

| Dataset | Micro-Expression Category | | |
|---|---|---|---|
| | *Negative* | *Positive* | *Surprise* |
| CASME II | 91 | 32 | 25 |
| SMIC (HS) | 70 | 51 | 43 |
| SMIC (VIS) | 28 | 23 | 20 |
| SMIC (NIR) | 28 | 23 | 20 |

*Repression* are categorized into *Negative*. The labels of *Surprise* samples are unchanged.

The new sample constitution information of SMIC and CASME II with respect to consistent micro-expression categorization is shown in Table 1. From Table 1, it can be seen that class imbalance problem exists in CASME II and SMIC (HS) datasets, which means the number of one type of micro-expression samples is significantly larger or lower than other types of micro-expression samples. Consequently, for better reporting the experimental results, we choose two metrics widely used in cross-database speech emotion recognition, i.e., unweighted average recall (**UAR**) and weighted average recall (**WAR**) [33], in the experiments. According to the definition in [33], **WAR** is the normal recognition accuracy, while **UAR** is the mean accuracy of each class divided by the number of classes without consideration of samples per class. It can comprehensively reveal the true performance of one method by comparing **WAR** and **UAR** of this method. For example, if there is a big gap between a high **WAR** and a low **UAR** in a method, it usually occurs that most of target samples are predicted by this method as the micro-expression category whose sample number percentage is dominant among all the micro-expression samples. Consequently, this method cannot be deemed to perform good even though it achieves a high **WAR** (recognition accuracy).

For comparison purpose, some recently proposed well-performing cross-database emotion (speech emotion and facial expression) recognition methods including KMM [13, 14], KLIEP [13, 35], uLSIF [13, 19], and STM [3, 4] are chosen. Since STM method has been introduced in related works section, we briefly introduce the other three methods here. KMM was proposed by Huang et al. [14] and aims at learning a set of weighted parameters for source samples such that weighted source samples and target samples satisfy the MMD criterion and hence they obey the same or feature distributions. KLIEP was proposed by Sugiyama et al. [35]. The key idea of KLIEP is to learn the importance weights for target samples based on KL divergence to balance the feature distribution gap between the source and target samples. uLSIF is also a method for learning the importance weights for source samples and was proposed by Kanamori et al. [19]. The most novelty of uLSIF is that it can be converted to a regularized model and has a closed form solution such that it is very fast compard with KMM. Note that SVM is served as the classifier for all the methods. Besides, SVM without any domain adaptation is also included for comparison. The detailed parameter setting of all the methods and micro-expression features are shown as follows:

(1) For micro-expression feature, we use uniform LBP-TOP [44] and the neighboring radius $R$ and the number of the neighboring points $P$ for LBP operator on three orthogonal planes are set as 3 and 8, respectively. Besides, following the work of [45], a multi-scale spatial division scheme consisting of $1 \times 1$, $2 \times 2$, $4 \times 4$, and $8 \times 8$ grids is adopted to partition the micro-expression video clips into a few facial blocks. Consequently, each micro-expression sample is described by a feature vector comprising the LBP-TOP vectors of all the facial blocks.

(2) We use linear kernel function and set $C = 1$ for SVM in the experiments. Meanwhile, for fair comparison, linear kernel function is adopted for all the methods throughout all the experiments.

(3) For KMM, according to the suggestion of [14], its two important parameters including the upper limit of importance weight $B$ and $\epsilon$ are set as 1000 and $\sqrt{n_{tr}} - 1/\sqrt{n_{tr}}$, where $n_{tr}$ denotes the number of training samples. For STM, following the work of [3, 4], the upper limit of importance weight $B$ and $\epsilon$ are set as the same values with those of KMM.

(4) For KLIEP, no parameter needs be set, while for uLSIF, STM, and the proposed TSRG, there are trade-off parameters to be set. Since the label information of target domain is entirely unknown, cross-validation method is not feasible for determining the trade-off parameters. Consequently, to offer a fair comparison among all the methods, in the experiments we use the parameter grid search strategy for these methods and report the best results which correspond to the optimal trade-off parameters. The optimal trade-off parameters of these three methods are as follows: (a) For uLSIF, its trade-off parameter $\lambda$ is fixed at 570, $36 \times 10^5$, 89, 1, 79, and $42 \times 10^5$ in Exp.1, Exp.2, Exp.3, Exp.4, Exp.5, and Exp.6, respectively. (b) For STM, its trade-off parameter $\lambda$ is fixed at 4, 0.04, 8, 0.02, 15, and 0.03 in Exp.1, Exp.2, Exp.3, Exp.4, Exp.5, and Exp.6, respectively. (c) For our TSRG, the optimal trade-off parameters $(\lambda, \mu)$ are (0.001, 0.03), (0.1, 0.007), (1, 0.07), (0.01, 0.005), (0.01, 0.003), and (10, 0.001) in Exp.1, Exp.2, Exp.3, Exp.4, Exp.5, and Exp.6, respectively.

### 4.2 Experimental Results and Analysis

The **WAR** and **UAR** of six experiments obtained by all the methods are depicted in Tables 2 and 3, respectively. From Tables 2 and 3, it is clear that in all the experiments, the proposed TSRG method has promising increases in the performance over SVM without domain adaptation. Moreover, our method achieves the best **WAR**, **UAR** or both in most cases, e.g., Exp.1 (best **WAR**), Exp.3 (best **UAR**) and Exp.4 (best both **WAR** and **UAR**). In addition, although STM performs better in term of **UAR** in Exp.1 and uLSIF outperforms the proposed TSRG in term of **WAR** in Exp.2 and in terms of both **WAR** and **UAR** in Exp.5, we can find that our TSRG is actually overall competitive against uLSIF and STM in these experiments.

Furthermore, it can be found that most of methods perform poorly in Exp.6, which indicates that Exp. 6 is a tough task for these methods. However, it is surprising to see that STM and the proposed TSRG achieve the **WAR** of 60.81% and 42.57%, respectively, which are considerably higher than other four methods in this experiment. By further comparing the **WAR** and **UAR** of STM and the proposed TSRG in this experiment, we can clearly find that the gap between **WAR** and **UAR** of the proposed TSRG (42.57% and 43.94%) is much narrower than STM (60.81% and 34.32%). It is very likely due to the extreme class imbalance problem exists in CASME II. As Table 1 shows, the percentage of *Negative* samples is dominant in CASME II. Because of this, most of CASME II samples may be mistakenly predicted as *Negative* by STM and hence the STM method achieves a low **UAR** although its **WAR** is leading among all the methods. In other words, the gap of our TSRG is actually more acceptable, which indicates that the proposed TSRG method is less affected by the extreme class imbalance problem exists in CASME II and is more applicable to this experiment.

Table 2: Experimental results on CASME II and one subset of SMIC (HS, VIS, NIR) databases in recognizing three micro-expressions (*Negative*, *Positive*, and *Surprise*). The results are reported in term of weighted average recall (WAR).

| # | Source Database | Target Database | SVM | KMM | KLIEP | uLSIF | STM | TSRG |
|---|---|---|---|---|---|---|---|---|
| Exp.1 | CASME II | SMIC (HS) | 42.07 | 37.20 | 45.73 | 47.56 | 46.34 | **50.61** |
| Exp.2 | SMIC (HS) | CASME II | 24.32 | 27.70 | 24.32 | **66.22** | 58.11 | 64.19 |
| Exp.3 | CASME II | SMIC (VIS) | 45.07 | 26.76 | 45.07 | 54.93 | 56.34 | **60.56** |
| Exp.4 | SMIC (VIS) | CASME II | 36.49 | 29.73 | 36.49 | 39.19 | 51.35 | **52.03** |
| Exp.5 | CASME II | SMIC (NIR) | 45.07 | 26.76 | 47.89 | **50.70** | 47.89 | 45.07 |
| Exp.6 | SMIC (NIR) | CASME II | 16.22 | 21.62 | 16.22 | 29.05 | **60.81** | 42.57 |

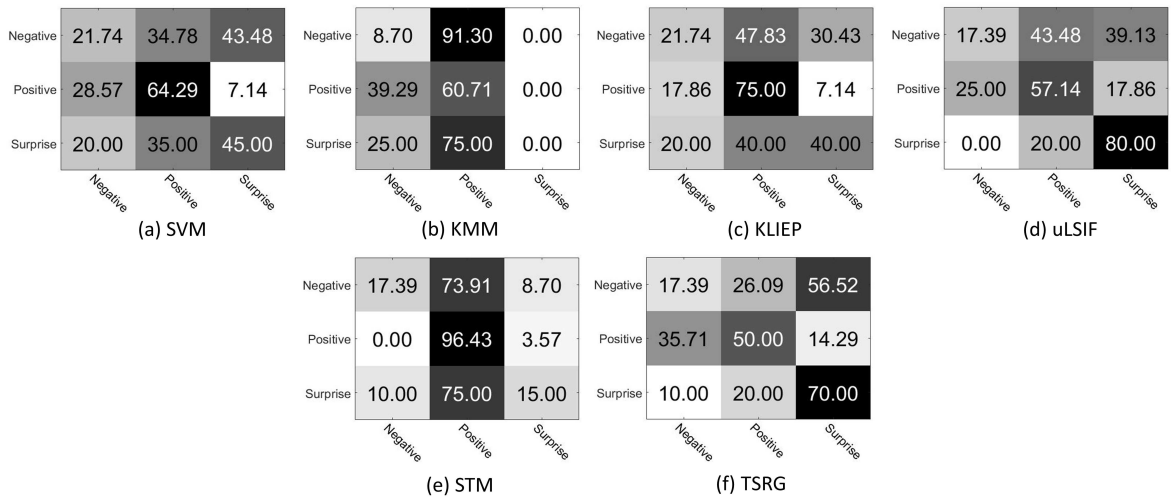

Figure 3: The confusion matrices of all the comparison methods in Exp.5. From (a) to (f), the experimental results correspond to SVM, KMM, KLIEP, STM, and TSRG, respectively.

In order to check the above analysis and further observe the interference of class imbalance in CASME II to each method, we select two experiments, i.e., Exp.5 and Exp.6, where CASME II is served as source and target database, respectively, as the representatives and draw the confusion matrices of all six methods in these two experiments. All the confusion matrices in Exp.5 and Exp.6 are shown in Figs. 3 (Exp.5) and 4 (Exp.6), respectively. From Figs. 3 and 4, some interesting findings and conclusions can be obtained:

(1) Firstly, as the confusion matrix of STM in Fig. 4 (Exp.6) shows, it is clear that nearly all the samples of CASME II database are predicted as *Negative* micro-expression by STM, which is consistent with our analysis previously. Consequently, the above analysis to explain why a big gap between the **WAR** and **UAR** exists in STM is reasonable.

(2) Secondly, it can be found that for all the methods, three micro-expressions are much more easily confused in the case when the CASME II is served as the target database (Fig. 4 and Exp.6) than the opposite case where the CASME II is used as the source database (Fig. 3 and Exp.5). This indicates that if the class imbalance problem occurred in the target database, domain adaptation methods would be more possibly interfered and hence their performance may be decreased.

(3) Thirdly, we notice that compared with Exp.6 (Fig. 4), the confusion among different micro-expressions in Exp.5 (Fig. 3) is relieved promisingly for all the methods. However, it should be pointed out that in Exp.5 (Fig. 3), the results of most methods are still unsatisfactory. Besides the class imbalance problem in CASME II, we think the heterogeneous problem of facial images between CASME II and SMIC (NIR) may be a major factor as well. From the works of [23, 43], we know that the samples of SMIC (NIR) and CASME II are recorded by a near-infrared camera and a high speed color camera, respectively. Consequently, the images of video clips from these two datasets are considerably heterogeneous and looks very different, which adds difficulty to the experiments between such two datasets.

Table 3: Experimental results on CASME II and one subset of SMIC (HS, VIS, NIR) databases in recognizing three micro-expressions (*Negative*, *Positive*, and *Surprise*). The results are reported in term of unweighted average recall (UAR).

| # | Source Database | Target Database | SVM | KMM | KLIEP | uLSIF | STM | TSRG |
|---|---|---|---|---|---|---|---|---|
| Exp.1 | CASME II | SMIC (HS) | 37.91 | 34.66 | 42.32 | 47.61 | **49.64** | 47.48 |
| Exp.2 | SMIC (HS) | CASME II | 35.14 | 39.67 | 35.14 | 43.43 | 41.17 | **49.58** |
| Exp.3 | CASME II | SMIC (VIS) | 42.67 | 25.72 | 41.93 | 53.12 | 53.10 | **60.15** |
| Exp.4 | SMIC (VIS) | CASME II | 48.94 | 44.02 | 49.91 | 53.31 | 39.99 | **56.03** |
| Exp.5 | CASME II | SMIC (NIR) | 43.67 | 23.14 | 45.58 | **51.51** | 42.94 | 45.80 |
| Exp.6 | SMIC (NIR) | CASME II | 31.42 | 36.54 | 31.71 | 40.26 | 34.32 | **43.94** |

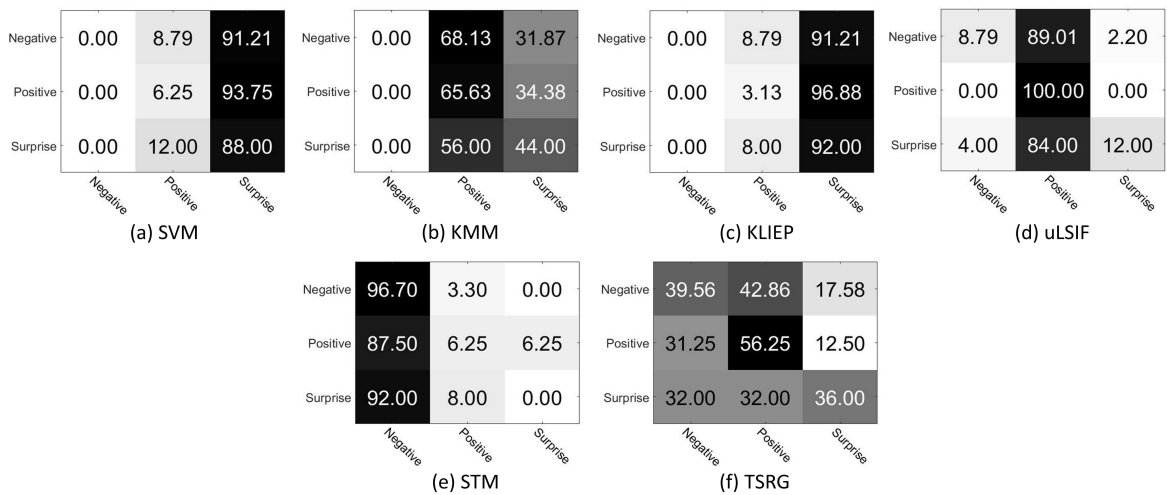

Figure 4: The confusion matrices of all the comparison methods in Exp.6. From (a) to (f), the experimental results correspond to SVM, KMM, KLIEP, STM, and TSRG, respectively.

(4) Finally, it is clear to see that even though most methods perform poorly in Exp.6 (Fig. 4), the proposed TSRG method can still achieve a satisfactory result, in which the extreme micro-expression confusion occurs in other comparison methods is promisingly alleviated (see Fig. 4 (f)).

## 5 CONCLUSION

In this paper, we have proposed a Target Sample Re-Generator (TSRG) method to deal with cross-database micro-expression recognition problem, which is more challenging than conventional micro-expression recognition one. By inputting source and target micro-expression samples, TSRG can re-generate the source and target samples, where the re-generated source samples are still the original ones, while the re-generated target samples would share the similar feature distribution as the source samples. Consequently, the classifier learned based on the source samples can accurately predict the micro-expression categories of the target samples. Extensive cross-database micro-expression recognition experiments between CASME II and SMIC databases are conducted to evaluate the performance of TSRG method. Experimental results show that TSRG method can achieve promising results and outperform lots of recent proposed state-of-the-art cross-database emotion recognition methods.

## 6 ACKNOWLEDGMENTS

This work was supported by the National Basic Research Program of China under Grant 2015CB351704, the National Natural Science Foundation of China under Grant 61231002 and Grant 61572009, China Scholarship Council, Academy of Finland, Tekes Fidipro Program and Infotech Oulu. We also gratefully acknowledge the support of NVIDIA Corporation with the donation of the Titan X Pascal GPU used for this research.

## REFERENCES
[1] Karsten M Borgwardt, Arthur Gretton, Malte J Rasch, Hans-Peter Kriegel, Bernhard Schölkopf, and Alex J Smola. 2006. Integrating structured biological data

by kernel maximum mean discrepancy. *Bioinformatics* 22, 14 (2006), e49–e57.
[2] George Casella and Roger L Berger. 2002. *Statistical inference*. Vol. 2. Duxbury Pacific Grove, CA.
[3] Wen-Sheng Chu, Fernando De la Torre, and Jeffery F Cohn. 2013. Selective transfer machine for personalized facial action unit detection. In *Proceedings of the IEEE Conference on Computer Vision and Pattern Recognition*. 3515–3522.
[4] Wen-Sheng Chu, Fernando De la Torre, and Jeffrey F Cohn. 2017. Selective transfer machine for personalized facial expression analysis. *IEEE Transactions on Pattern Analysis and Machine Intelligence* 39, 3 (2017), 529–545.
[5] Jun Deng, Xinzhou Xu, Zixing Zhang, Sascha Frühholz, and Björn Schuller. 2017. Universum Autoencoder-Based Domain Adaptation for Speech Emotion Recognition. *IEEE Signal Processing Letters* 24, 4 (2017), 500–504.
[6] Jun Deng, Zixing Zhang, Florian Eyben, and Bjorn Schuller. 2014. Autoencoder-based unsupervised domain adaptation for speech emotion recognition. *IEEE Signal Processing Letters* 21, 9 (2014), 1068–1072.
[7] Jun Deng, Zixing Zhang, Erik Marchi, and Bjorn Schuller. 2013. Sparse autoencoder-based feature transfer learning for speech emotion recognition. In *2013 Humaine Association Conference on Affective Computing and Intelligent Interaction (ACII)*. IEEE, 511–516.
[8] Richard O Duda, Peter E Hart, and David G Stork. 2012. *Pattern classification*. John Wiley & Sons.
[9] Paul Ekman and Wallace V Friesen. 1977. Facial action coding system. (1977).
[10] MG Frank, Malgorzata Herbasz, Kang Sinuk, A Keller, and Courtney Nolan. 2009. I see how you feel: Training laypeople and professionals to recognize fleeting emotions. In *The Annual Meeting of the International Communication Association. Sheraton New York, New York City*.
[11] Mark G Frank, Carl J Maccario, and Venugopal Govindaraju. 2009. Behavior and security. *Protecting Airline Passengers in the Age of Terrorism. Greenwood Pub Group, Santa Barbara, California* (2009), 86–106.
[12] Felix A Gers, Nicol N Schraudolph, and Jürgen Schmidhuber. 2002. Learning precise timing with LSTM recurrent networks. *Journal of Machine Learning Research* 3, Aug (2002), 115–143.
[13] Ali Hassan, Robert Damper, and Mahesan Niranjan. 2013. On acoustic emotion recognition: compensating for covariate shift. *IEEE Transactions on Audio, Speech, and Language Processing* 21, 7 (2013), 1458–1468.
[14] Jiayuan Huang, Arthur Gretton, Karsten M Borgwardt, Bernhard Schölkopf, and Alex J Smola. 2006. Correcting sample selection bias by unlabeled data. In *Advances in Neural Information Processing Systems*. 601–608.
[15] Xiaohua Huang, Su-Jing Wang, Guoying Zhao, and Matti Pietikäinen. 2015. Facial micro-expression recognition using spatiotemporal local binary pattern with integral projection. In *Proceedings of the IEEE International Conference on Computer Vision Workshops*. 1–9.
[16] Xiaohua Huang, Guoying Zhao, Xiaopeng Hong, Wenming Zheng, and Matti Pietikäinen. 2016. Spontaneous facial micro-expression analysis using spatiotemporal completed local quantized patterns. *Neurocomputing* 175 (2016), 564–578.
[17] Zhengwei Huang, Wentao Xue, Qirong Mao, and Yongzhao Zhan. 2016. Unsupervised domain adaptation for speech emotion recognition using PCANet. *Multimedia Tools and Applications* (2016), 1–15.
[18] Shuiwang Ji and Jieping Ye. 2009. An accelerated gradient method for trace norm minimization. In *Proceedings of the 26th annual International Conference on Machine Learning*. ACM, 457–464.
[19] Takafumi Kanamori, Shohei Hido, and Masashi Sugiyama. 2009. A least-squares approach to direct importance estimation. *The Journal of Machine Learning Research* 10 (2009), 1391–1445.
[20] Dae Hoe Kim, Wissam J Baddar, and Yong Man Ro. 2016. Micro-Expression Recognition with Expression-State Constrained Spatio-Temporal Feature Representations. In *Proceedings of the 2016 ACM on Multimedia Conference*. ACM, 382–386.
[21] Alex Krizhevsky, Ilya Sutskever, and Geoffrey E Hinton. 2012. Imagenet classification with deep convolutional neural networks. In *Advances in Neural Information Processing Systems*. 1097–1105.
[22] Xiaobai Li, Xiaopeng Hong, Antti Moilanen, Xiaohua Huang, Tomas Pfister, Guoying Zhao, and Matti Pietikäinen. 2017. Towards Reading Hidden Emotions: A Comparative Study of Spontaneous Micro-expression Spotting and Recognition Methods. *IEEE Transactions on Affective Computing* (2017).
[23] Xiaobai Li, Tomas Pfister, Xiaohua Huang, Guoying Zhao, and Matti Pietikäinen. 2013. A spontaneous micro-expression database: Inducement, collection and baseline. In *Proceedings of the 10th IEEE International Conference and Workshops on Automatic Face and Gesture Recognition (FG)*. IEEE, 1–6.
[24] Zhouchen Lin, Minming Chen, and Yi Ma. 2010. The augmented lagrange multiplier method for exact recovery of corrupted low-rank matrices. *arXiv preprint arXiv:1009.5055* (2010).
[25] Yong-Jin Liu, Jin-Kai Zhang, Wen-Jing Yan, Su-Jing Wang, Guoying Zhao, and Xiaolan Fu. 2016. A Main Directional Mean Optical Flow Feature for Spontaneous Micro-Expression Recognition. *IEEE Transactions on Affective Computing* 7, 4 (2016), 299–310.
[26] Ping Lu, Wenming Zheng, Ziyan Wang, Qiang Li, Yuan Zong, Minghai Xin, and Lenan Wu. 2016. Micro-Expression Recognition by Regression Model and Group Sparse Spatio-Temporal Feature Learning. *IEICE TRANSACTIONS on Information and Systems* 99, 6 (2016), 1694–1697.
[27] Maureen OfiSullivan, Mark G Frank, Carolyn M Hurley, and Jaspreet Tiwana. 2009. Police lie detection accuracy: The effect of lie scenario. *Law and Human Behavior* 33, 6 (2009), 530–538.
[28] Sinno Jialin Pan and Qiang Yang. 2010. A survey on transfer learning. *IEEE Transactions on Knowledge and Data Engineering* 22, 10 (2010), 1345–1359.
[29] Tomas Pfister, Xiaobai Li, Guoying Zhao, and Matti Pietikäinen. 2011. Recognising spontaneous facial micro-expressions. In *International Conference on Computer Vision*. IEEE, 1449–1456.
[30] John A Ruiz-Hernandez and Matti Pietikäinen. 2013. Encoding local binary patterns using the re-parametrization of the second order gaussian jet. In *Proceedings of the 10th IEEE International Conference and Workshops on Automatic Face and Gesture Recognition (FG)*. IEEE, 1–6.
[31] Hesam Sagha, Jun Deng, Maryna Gavryukova, Jing Han, and Björn Schuller. 2016. Cross lingual speech emotion recognition using canonical correlation analysis on principal component subspace. In *IEEE International Conference on Acoustics, Speech and Signal Processing (ICASSP)*. IEEE, 5800–5804.
[32] Enver Sangineto, Gloria Zen, Elisa Ricci, and Nicu Sebe. 2014. We are not all equal: Personalizing models for facial expression analysis with transductive parameter transfer. In *Proceedings of the 22nd ACM International Conference on Multimedia*. ACM, 357–366.
[33] Bjorn Schuller, Bogdan Vlasenko, Florian Eyben, Martin Wollmer, Andre Stuhlsatz, Andreas Wendemuth, and Gerhard Rigoll. 2010. Cross-corpus acoustic emotion recognition: Variances and strategies. *IEEE Transactions on Affective Computing* 1, 2 (2010), 119–131.
[34] Peng Song, Wenming Zheng, Shifeng Ou, Xinran Zhang, Yun Jin, Jinglei Liu, and Yanwei Yu. 2016. Cross-corpus speech emotion recognition based on transfer non-negative matrix factorization. *Speech Communication* 83 (2016), 34–41.
[35] Masashi Sugiyama, Shinichi Nakajima, Hisashi Kashima, Paul V Buenau, and Motoaki Kawanabe. 2008. Direct importance estimation with model selection and its application to covariate shift adaptation. In *Advances in Neural Information Processing Systems*. 1433–1440.
[36] Su-Jing Wang, Wen-Jing Yan, Xiaobai Li, Guoying Zhao, and Xiaolan Fu. 2014. Micro-expression recognition using dynamic textures on tensor independent color space. In *22nd International Conference on Pattern Recognition (ICPR)*. IEEE, 4678–4683.
[37] Su-Jing Wang, Wen-Jing Yan, Xiaobai Li, Guoying Zhao, Chun-Guang Zhou, Xiaolan Fu, Minghao Yang, and Jianhua Tao. 2015. Micro-expression recognition using color spaces. *IEEE Transactions on Image Processing* 24, 12 (2015), 6034–6047.
[38] Su-Jing Wang, Wen-Jing Yan, Guoying Zhao, Xiaolan Fu, and Chun-Guang Zhou. 2014. Micro-expression recognition using robust principal component analysis and local spatiotemporal directional features. In *Workshop at the European Conference on Computer Vision*. Springer, 325–338.
[39] Yandan Wang, John See, Raphael C-W Phan, and Yee-Hui Oh. 2014. Lbp with six intersection points: Reducing redundant information in lbp-top for micro-expression recognition. In *Asian Conference on Computer Vision*. Springer, 525–537.
[40] John Wright, Arvind Ganesh, Shankar Rao, Yigang Peng, and Yi Ma. 2009. Robust principal component analysis: Exact recovery of corrupted low-rank matrices via convex optimization. In *Advances in Neural Information Processing Systems*. 2080–2088.
[41] Feng Xu, Junping Zhang, and James Z Wang. 2017. Microexpression identification and categorization using a facial dynamics map. *IEEE Transactions on Affective Computing* 8, 2 (2017), 254–267.
[42] Keyu Yan, Wenming Zheng, Zhen Cui, and Yuan Zong. 2016. Cross-Database Facial Expression Recognition via Unsupervised Domain Adaptive Dictionary Learning. In *International Conference on Neural Information Processing*. Springer, 427–434.
[43] Wen-Jing Yan, Xiaobai Li, Su-Jing Wang, Guoying Zhao, Yong-Jin Liu, Yu-Hsin Chen, and Xiaolan Fu. 2014. CASME II: An improved spontaneous micro-expression database and the baseline evaluation. *PloS one* 9, 1 (2014), e86041.
[44] Guoying Zhao and Matti Pietikäinen. 2007. Dynamic texture recognition using local binary patterns with an application to facial expressions. *IEEE Transactions on Pattern Analysis and Machine Intelligence* 29, 6 (2007), 915–928.
[45] Guoying Zhao and Matti Pietikäinen. 2009. Boosted multi-resolution spatiotemporal descriptors for facial expression recognition. *Pattern Recognition Letters* 30, 12 (2009), 1117–1127.
[46] Wenming Zheng and Xiaoyan Zhou. 2015. Cross-pose color facial expression recognition using transductive transfer linear discriminat analysis. In *IEEE International Conference on Image Processing (ICIP)*. IEEE, 1935–1939.
[47] Wenming Zheng, Yuan Zong, Xiaoyan Zhou, and Minghai Xin. 2016. Cross-Domain Color Facial Expression Recognition Using Transductive Transfer Subspace Learning. *IEEE Transactions on Affective Computing* (2016).
[48] Yuan Zong, Wenming Zheng, Tong Zhang, and Xiaohua Huang. 2016. Cross-corpus speech emotion recognition based on domain-adaptive least-squares regression. *IEEE Signal Processing Letters* 23, 5 (2016), 585–589.